\documentclass{article}
\usepackage{spconf,amsmath,graphicx}

\usepackage{enumitem}
\usepackage{amsmath}

\usepackage{textcomp}
\usepackage[export]{adjustbox}
\usepackage{amsmath,graphicx}
\usepackage[utf8]{inputenc}
\usepackage[colorinlistoftodos]{todonotes}
\usepackage{tabularx}
\usepackage{multirow}
\usepackage{interval}
\usepackage[lofdepth,lotdepth]{subfig}
\newcolumntype{L}[1]{>{\raggedright\let\newline\\\arraybackslash\hspace{0pt}}m{#1}}
\newcolumntype{C}[1]{>{\centering\let\newline\\\arraybackslash\hspace{0pt}}m{#1}}
\usepackage[colorlinks=true,citecolor=blue]{hyperref}
\usepackage{siunitx}
\usepackage{placeins}
\usepackage{graphicx}
\usepackage{multicol}
\usepackage{float}
\restylefloat{table}
\usepackage{hhline}
\usepackage{cite}
\usepackage{dsfont}
\usepackage{amssymb}
\newlength{\tempdima}
\newcommand{\rowname}[1]% #1 = text
{\rotatebox{90}{\makebox[\tempdima][c]{#1}}}
\setlist{nosep, leftmargin=14pt}

\usepackage{mwe} % to get dummy images

\title{Self Adversarial Attack as an Augmentation Method for Immunohistochemical Stainings}
\name{Jelica Vasiljević$^{\star,\dagger,\text{\textbardbl}}$ \quad Friedrich Feuerhake$^{\text{\textdaggerdbl},\maltese}$ \quad C\'{e}dric Wemmert$^{\star}$  \quad Thomas Lampert$^{\star}$}
\address{$^{\star}$ICube, University of Strasbourg, France \quad 
		 $^{\dagger}$University of Belgrade, Serbia \\
		 $^{\text{\textbardbl}}$Faculty of Science, University of Kragujevac, Serbia \\ $^{\text{\textdaggerdbl}}$ Institute of Pathology, Hannover Medical School, Germany \quad $^{\maltese}$ University Clinic, Freiburg, Germany}

\begin{document}

\maketitle

\begin{abstract}
It has been shown that unpaired image-to-image translation methods constrained by cycle-consistency hide the information necessary for accurate input reconstruction as imperceptible noise. We demonstrate that, when applied to histopathology data, this hidden noise appears to be related to stain specific features and show that this is the case with two immunohistochemical stainings during translation to \textit{Periodic acid-Schiff (PAS)}, a histochemical staining method commonly applied in renal pathology. 
Moreover, by perturbing this hidden information, the translation models produce different, plausible outputs. We demonstrate that this property can be used as an augmentation method which, in a case of supervised glomeruli segmentation, leads to improved performance.
\end{abstract}
\section{Introduction}
\label{sec:intro}

One of the greatest obstacles for the effective application of deep learning techniques to digital pathology is the shortage of high-quality annotated data.
The annotation process itself is time consuming and expensive 
as expert domain knowledge is required for most complex annotations and alternative approaches such as crowd sourcing are limited by the need of specific task design and intensive training \cite{grote2018crowdsourcing}.
The problem is complicated by tissue appearance variability, which can occur due to different stainings, patients, procedures between different laboratories, and/or the microscope and imaging device \cite{Leo2016EvaluatingSO}. All of this imposes a domain shift to which deep models are very sensitive \cite{csurka2017comprehensive}, making their application difficult in clinical practice.

Due to their ability to produce high quality visual outputs, Generative Adversarial Networks (GANs) \cite{goodfellow2014generative} have recently been applied to medical imaging in general and digital pathology. Finding use in histopathology to reduce intra-stain variance \cite{Tellez2019QuantifyingTE}; for virtual staining \cite{Mercan2020VirtualSF, Lahiani2018VirtualizationOT}; and for augmentation \cite{Vasiljevic20,PathologyGAN20}. Virtual staining has shown that an unpaired image-to-image translation GAN is able to translate between stains. The same tissue can be (artificially) stained in multiple stainings, which is hard (or even impossible) in realty \cite{Mercan2020VirtualSF}. CycleGAN is the most popular and promising unpaired image-to-image translation approach \cite{CycleGAN2017, gadermayr2018which}. Nevertheless, the less obvious limitations of such methods are rarely addressed in the medical imaging literature \cite{Mercan2020VirtualSF}. For example, such models produce realistic translations between very different stains, which leads to the question: how is the model able to place stain related markers that are not present in the original stain? This article moves towards answering this question.

\begin{figure*}[tb!]
	%\begin{minipage}[b]{0.1\linewidth}
	\centering
	\includegraphics[width=0.8\linewidth]{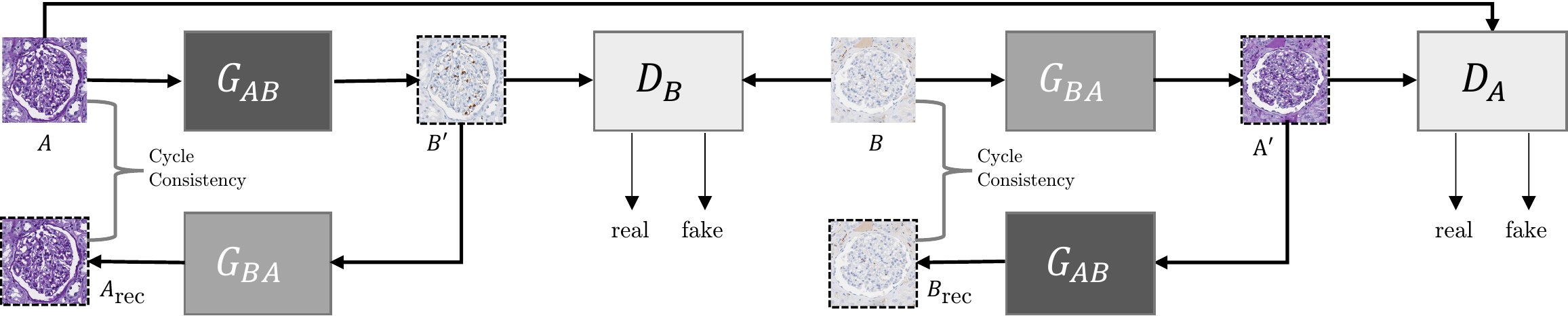}
	%\end{minipage}
	\caption{CycleGAN approach (with PAS and CD68 staining examples). Framed images are translated, i.e.\ `fake'.}
	\label{fig:cyclegan_method}
\end{figure*}

The computer vision community has recently shown with natural images that the cycle-consistency of CycleGANs renders them prone to self-adversarial attack \cite{Bashkirova19}. The CycleGAN (Fig.\ \ref{fig:cyclegan_method}) is composed of two translators:  one from staining A to B, $G_{AB}$, and another from B to A, $G_{BA}$. The cycle consistency enforces that the output of $G_{BA}$ matches the input of $G_{AB}$. To achieve this, each translator is forced to hide imperceptible information in its output. Our first contribution is to show that the hidden noise has a specific meaning in histopathology - it encodes stain-related markers. By perturbing this hidden noise, differently positioned stain-related markers are produced in the translated image (leaving the underlying tissue structure untouched).

This is exploited to introduce a new augmentation technique that increases the variability of stain-specific markers in histopathological data, with the goal of increasing a model's robustness when trained for non-stain-related tasks. 
We show that this increases the generalisation performance of a supervised deep learning approach for glomeruli segmentation, which forms this article's second contribution.

\begin{figure*}[tb]
    \centering
    \newcommand{\figheight}{0.12}
	\settoheight{\tempdima}{\includegraphics[width=\figheight\textwidth]{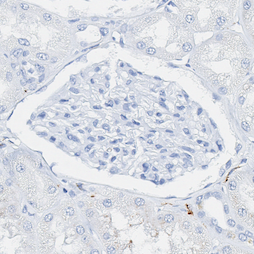}}%
	\begin{tabular}{c@{ }c@{ }c@{ }c@{ }c@{ }c@{ }c@{ }%c@{ }c@{ }c@{ }c@{ }
	}
	    \small{Original} & \multicolumn{5}{c}{\small{Reconstruction}}\\
	     &  \small{No Noise} & \small{$\sigma = 0.05$} & 
		%$\sigma = 0.075$ & 
		\small{$\sigma = 0.1$} & 
		\small{$\sigma = 0.2$} & 
		\small{$\sigma = 0.3$} & 
		%$\sigma = 0.5$ & $\sigma = 0.7$ & 
		\small{$\sigma = 0.9$}
		\\
		\rowname{\small{CD68}} \includegraphics[width=\figheight\textwidth]{img/noise/IFTA_Nx_0010_16_glomeruli_patch_0_orig.png}&
		%\rowname{Reconstruction}
		%\multirow{3}{*}{\rotatebox{90}{Reconstructions}}
		%&
		\includegraphics[width=\figheight\textwidth]{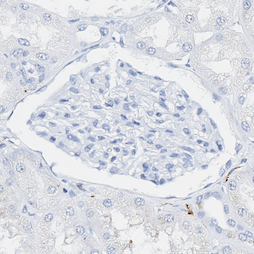} &
		\includegraphics[width=\figheight\textwidth]{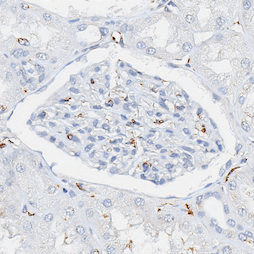} &
		\includegraphics[width=\figheight\textwidth]{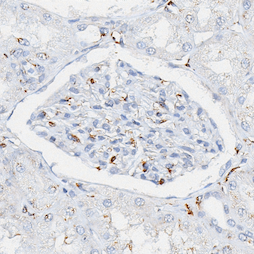} &
		\includegraphics[width=\figheight\textwidth]{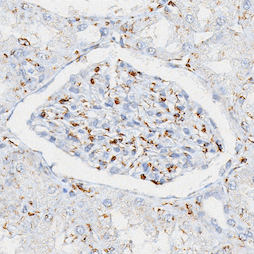} &
		\includegraphics[width=\figheight\textwidth]{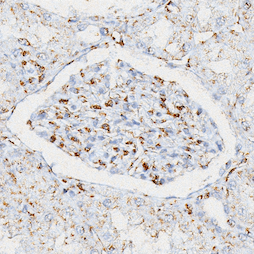} &
		\includegraphics[width=\figheight\textwidth]{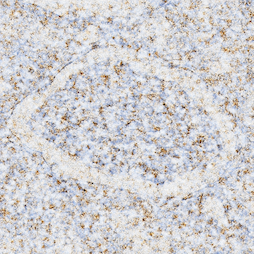}
		\\
		%\hline
		%\vspace{-1.75ex}
		%\\
		\rowname{\small{CD34}} \includegraphics[width=\figheight\textwidth]{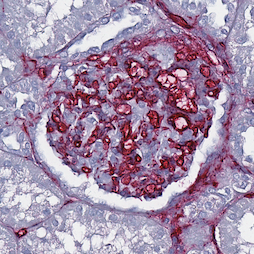}&
		%\rowname{Reconstruction}
		\includegraphics[width=\figheight\textwidth]{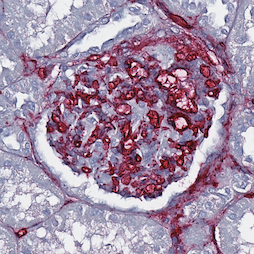}&
		\includegraphics[width=\figheight\textwidth]{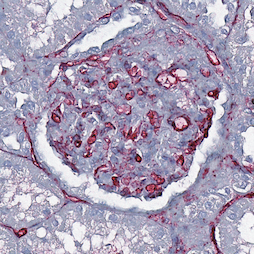}&
		\includegraphics[width=\figheight\textwidth]{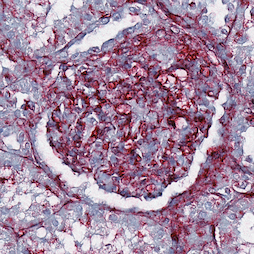}&
		\includegraphics[width=\figheight\textwidth]{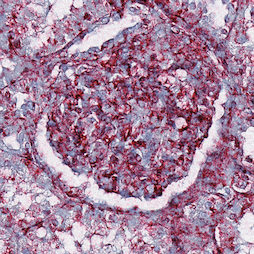} &
		\includegraphics[width=\figheight\textwidth]{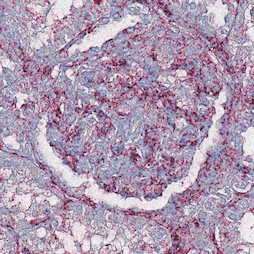}&
		\includegraphics[width=\figheight\textwidth]{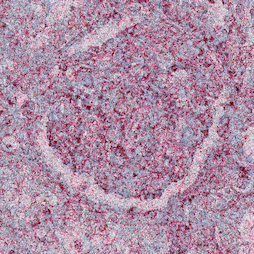}
	\end{tabular}%
	\caption{Generating variation by adding noise, the images are reconstructions of CD68/CD34 $\rightarrow$ PAS + $\mathcal{N}(0,\sigma)\rightarrow$ CD68/CD34\color{black}.}
	\label{fig:translation_added_noise}
\end{figure*}

\begin{figure*}[tb]
    \centering
    \newcommand{\figheight}{0.12}
	\settoheight{\tempdima}{\includegraphics[width=\figheight\textwidth]{img/noise/IFTA_Nx_0010_16_glomeruli_patch_0_orig.png}}%
	\begin{tabular}{c@{ } c@{ }c@{ }c@{ } c@{ }c@{ }c@{ }}
	    \small{Original} & \multicolumn{3}{c}{\small{$\sigma = 0.05$}}  & \multicolumn{3}{c}{\small{$\sigma = 0.1$}}\\
\rowname{\small{CD68}} \includegraphics[width=\figheight\textwidth]{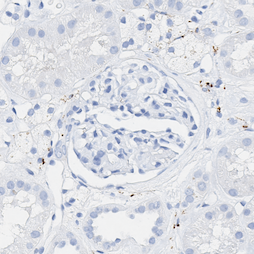} \hspace{1ex} &
\includegraphics[width=\figheight\textwidth]{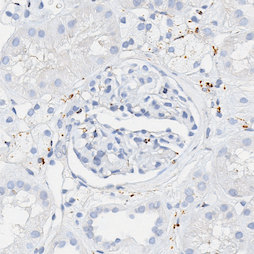}&
\includegraphics[width=\figheight\textwidth]{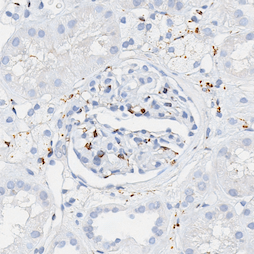}&
\includegraphics[width=\figheight\textwidth]{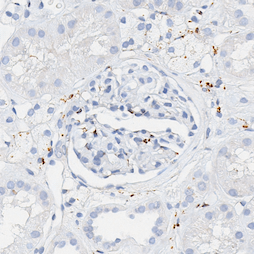} \hspace{1ex} &
\includegraphics[width=\figheight\textwidth]{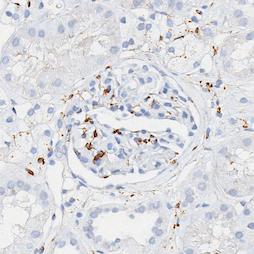}&
\includegraphics[width=\figheight\textwidth]{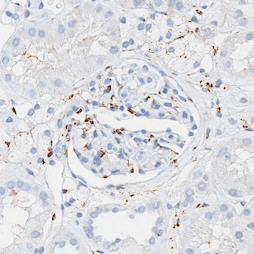}&
\includegraphics[width=\figheight\textwidth]{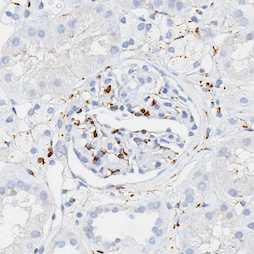}
		\\
\rowname{\small{CD34}} \includegraphics[width=\figheight\textwidth]{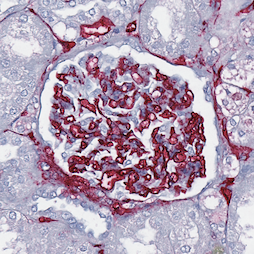} \hspace{1ex} &
\includegraphics[width=\figheight\textwidth]{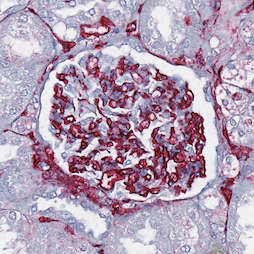} &
\includegraphics[width=\figheight\textwidth]{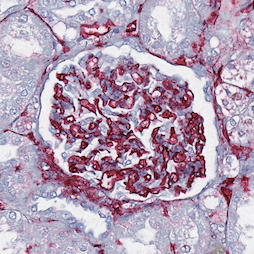} &
\includegraphics[width=\figheight\textwidth]{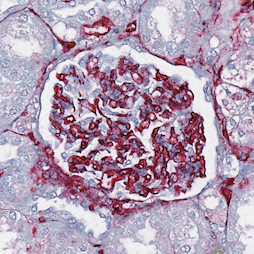} \hspace{1ex} &
\includegraphics[width=\figheight\textwidth]{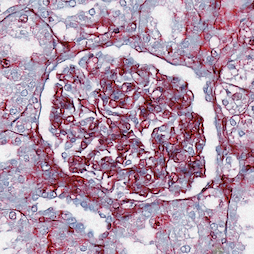} &
\includegraphics[width=\figheight\textwidth]{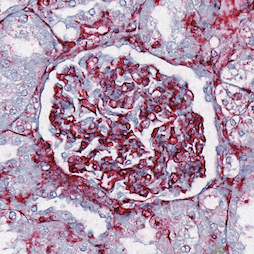} &
\includegraphics[width=\figheight\textwidth]{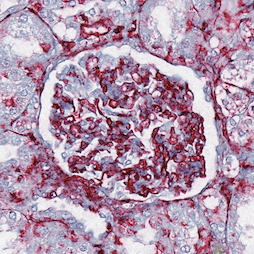} 
	\end{tabular}%
	\caption{Effects of additive Gaussian noise with the same standard deviation, the images are reconstructions of CD68/CD34 $\rightarrow$ PAS + $\mathcal{N}(0,\sigma)\rightarrow$ CD68/CD34}
	\label{fig:translation_added_same_std_noise}
\end{figure*}

We explore the mapping between Periodic acid-Schiff (PAS), a routine staining in renal pathology that is applied for general diagnostic purposes, and two immunohistochemical stainings (CD68 for macrophages and CD34 for blood vessel endothelium), which are performed for research or specific diagnostic purposes. Separate CycleGAN models are trained to translate between PAS stained tissue patches and each of the immunohistochemical stainings.

Section \ref{sec:adverseattack} of this article presents adversarial attacks in stain transfer; Section \ref{sec:augmentation} presents the new augmentation method and its evaluation; and Section \ref{sec:conclusions} our conclusions.

\section{Stain Transfer Self Adversarial Attack}
\label{sec:adverseattack}

\begin{figure*}[tb!]
	%\begin{minipage}[b]{0.1\linewidth}
	\centering
	\includegraphics[width=0.81\linewidth]{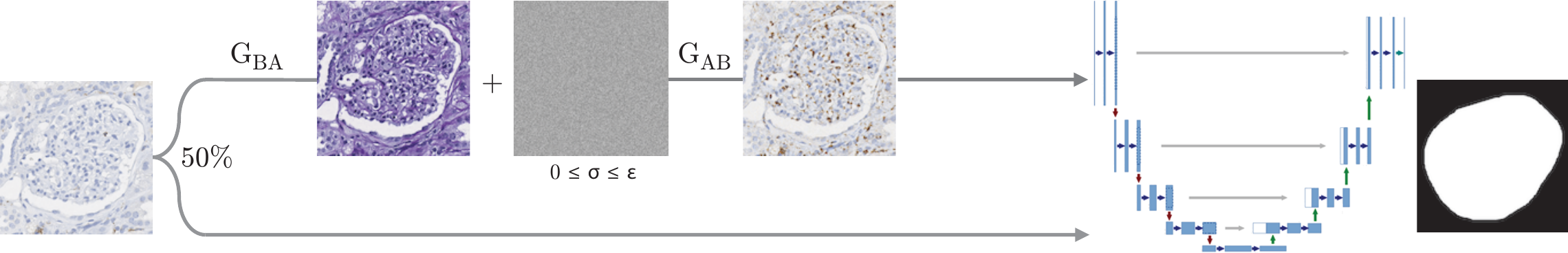}
	%\end{minipage}
	\caption{Proposed augmentation approach.}
	\label{fig:aug_approach}
\end{figure*}

Given samples of two histopathological stains $a \sim A$ and $b \sim B$, the goal is to learn two mappings (translators) $G_{AB}:a\sim A\rightarrow b\sim B$ and $G_{BA}:b \sim B \rightarrow a \sim A$. In order to do so, two adversarial discriminators $D_A$ and $D_B$ are jointly trained to distinguish between translated and real samples, i.e.\ $D_A$ aims to distinguish between real samples $a \sim A$ and $B$ translated to $A$ ($a'=G_{BA}(b), b \sim B$), while $D_B$ performs the equivalent task for $b \sim B$ and $b'=G_{AB}(a), a \sim A$. In addition to the adversarial loss \cite{goodfellow2014generative, CycleGAN2017}, the learning process is regularised by a cycle-consistency loss $\mathcal{L}_{cyc}$ that forces the generators to be consistent with each other \cite{CycleGAN2017}, such that
\begin{align}
\mathcal{L}_{cyc}(G_{AB},G_{BA}) = \mathds{E}_{a \sim A} [\| G_{BA}(G_{AB}(a))-a \|_1]\notag\\+ \mathds{E}_{b \sim B} [\| G_{AB}(G_{BA}(b))-b \|_1].
\label{eq:cycle_consistency}
\end{align}

In addition to the Haematoxylin counterstain (common to all the stainings studied herein) that highlights cell nuclei, CD68 marks a protein exclusively produced by macrophages, and CD34 stains a protein specific to the endothelial cells of blood vessels. PAS, as a chemical reaction staining glycolysated proteins in general, can highlight some parts of macrophages (co-located but not overlapping with CD68), the basal lamina of blood vessels (co-located with CD34), and other structures not highlighted by either CD68 nor CD34 that contain glycolysated proteins.
During translation from PAS to CD68, the model could choose not to produce macrophages (which would be a valid CD68 sample) but $D_{\text{CD68}}$ would easily discriminate real/fake images based on this absence, and therefore the model is biased to deduce their position from information present in PAS. Conversely, i.e.\ CD68 $\rightarrow$ PAS, the model should induce the presence of glycolysated proteins, for which CD68 is not specific. As such, the translation process is a many-to-many mapping (equivalent arguments can be made for PAS $\leftrightarrow$ CD34).

The cycle-consistency constraint Eq.\ \eqref{eq:cycle_consistency}, Fig.\ \ref{fig:cyclegan_method} forces compositions of translations ($A \rightarrow B \rightarrow A$) to accurately reconstruct the input.  Taking $\text{CD68}\rightarrow \text{PAS}\rightarrow\text{CD68}$ for example, macrophages in the reconstructed image should be in the same locations as those in the original, which implies that the intermediate PAS image contains additional information defining these macrophage positions. Bashkirova et al.\ \cite{Bashkirova19} recently showed that information necessary for perfect reconstruction takes the form of imperceptible low amplitude, high frequency noise in order to fool the discriminator, and recent literature \cite{Bashkirova19,Chu2017CycleGANAM} names this a self-adversarial attack. Since PAS does not contain information specific to macrophages/blood vessels this is likely to be the case.

\subsection{Dataset}
Tissue samples were collected from a cohort of $10$ patients who underwent tumor nephrectomy due to renal carcinoma. The kidney tissue was selected as distant as possible from the tumors to display largely normal renal glomeruli, some samples included variable degrees of pathological changes such as full or partial replacement of the functional tissue by fibrotic changes (``scerosis'') reflecting normal age-related changes or the renal consequences of general cardiovascular comorbidity (e.g.\ cardial arrhythmia, hypertension, arteriosclerosis).  
The paraffin-embedded samples were cut into \SI{3}{\micro\metre} thick sections and stained with either PAS or immunohistochemistry markers CD34 and CD68 using an automated staining instrument (Ventana Benchmark Ultra). Whole slide images (WSIs) were acquired using an Aperio AT2 scanner at $40{\times}\!$ magnification (a resolution of \SI[per-mode=fraction]{0.253}{\micro\metre\per{pixel}}). All glomeruli (healthy, partially sclerotic, and completely sclerotic) in each WSI were annotated and validated by pathology experts using Cytomine \cite{Cytomine16}. The dataset was divided into $4$ training, $2$ validation, and $4$ test patients.

For CycleGAN training, $5000$ random $508\times508$ pixel patches were extracted from the training patients and scaled to the range $[-1,1]$. The model's architecture (9 ResNet blocks) and training details were taken from the original article \cite{CycleGAN2017}.

\begin{table*}[t]
		\centering
	    \small{
		\begin{tabular}{L{0.8cm}| L{1.5cm} C{1.7cm} C{1.7cm}  C{1.7cm} | C{1.7cm} C{1.7cm} C{1.7cm}}
			%\hline \\[-2ex]
			\multirow{2}{*}{Stain} &
			 &
			\multicolumn{3}{c}{Baseline}& \multicolumn{3}{c}{Noise Augmented}\\
			& & F$_1$ & Precision & Recall & F$_1$ & Precision & Recall \\
			\hhline{========}
			\multirow{4}{*}{\shortstack{CD68}} & 10\% - 53 & 0.739 \footnotesize{(0.018)} & 0.754 \footnotesize{(0.047)} & \textbf{0.728} \footnotesize{(0.034)} & \textbf{0.767} \footnotesize{(0.036)} & \textbf{0.832} \footnotesize{(0.053)} & 0.713 \footnotesize{(0.044)} \\
			& 30\% - 159 & 0.812 \footnotesize{(0.026)} & 0.839 \footnotesize{(0.038)} & 0.788 \footnotesize{(0.038)} & \textbf{0.828} \footnotesize{(0.026)} & \textbf{0.848} \footnotesize{(0.065)} & \textbf{0.812} \footnotesize{(0.017)} \\
			& 60\% - 317 & 0.831 \footnotesize{(0.024)} & 0.812 \footnotesize{(0.037)} & \textbf{0.852} \footnotesize{(0.014)} &\textbf{0.856} \footnotesize{(0.017)} & \textbf{0.888} \footnotesize{(0.026)} & 0.826 \footnotesize{(0.021)} \\
			& 100\% - 529 & 0.853 \footnotesize{(0.018)} & 0.849 \footnotesize{(0.024)} & \textbf{0.858} \footnotesize{(0.020)} & \textbf{0.878} \footnotesize{(0.007)} & \textbf{0.899} \footnotesize{(0.023)} & \textbf{0.858} \footnotesize{(0.010)} \\
			\hline
			\multirow{4}{*}{\shortstack{CD34}} & 10\% - 57 & 0.837 \footnotesize{(0.017)} & 0.770 \footnotesize{(0.033)} & \textbf{0.919} \footnotesize{(0.009)} & \textbf{0.839} \footnotesize{(0.035)} & \textbf{0.778} \footnotesize{(0.061)} & 0.913 \footnotesize{(0.008)} \\
			& 30\% - 170 & 0.877 \footnotesize{(0.012)} & 0.841 \footnotesize{(0.030)} & \textbf{0.917} \footnotesize{(0.012)} & \textbf{0.890} \footnotesize{(0.011)} & \textbf{0.867} \footnotesize{(0.023)} & 0.916 \footnotesize{(0.009)} \\
			& 60\% - 341 & 0.882 \footnotesize{(0.008)} & 0.840 \footnotesize{(0.015)} & \textbf{0.927} \footnotesize{(0.005)} & \textbf{0.901} \footnotesize{(0.007)} & \textbf{0.884} \footnotesize{(0.019)} & 0.919 \footnotesize{(0.010)} \\
			& 100\% - 568 & 0.888 \footnotesize{(0.015)} & 0.849 \footnotesize{(0.033)} & \textbf{0.931} \footnotesize{(0.010)} & \textbf{0.903} \footnotesize{(0.006)} & \textbf{0.888} \footnotesize{(0.014)} & 0.919 \footnotesize{(0.009)} \\
			\hline
		\end{tabular}
	}
	\caption{Quantitative results, standard deviations are in parentheses, \# of glomeruli training patches follow the data percentages.}
	\label{tab: results}
\end{table*}

\subsection{Results and Analysis}

Figure \ref{fig:translation_added_noise} shows that translation output (i.e.\ reconstructed input, $B_\text{rec}$) variance is directly proportional to the level of additive noise and Fig.\ \ref{fig:translation_added_same_std_noise} shows that different translations result from varying noise of the same standard deviation.

As such, they give evidence to support that when translating between immunohistochemical and histochemical stains, imperceptible noise is present in the intermediate translation and this contains information about stain-related markers (this is related to macrophages marked in brown, and blood vessel endothelium marked in red in CD68 and CD34 respectively). Thus, changing the encoded noise changes the reconstruction of stain related markers. This noise can be perturbed by introducing additive zero-mean Gaussian noise to the intermediate translation \cite{Bashkirova19}. The amount of stain related characteristics can be controlled through the Gaussian's standard deviation. The physical accuracy of the resulting stain-related markers remains an open question, but the fact that they are positioned in plausible locations opens the possibility of exploiting them to reduce a model's sensitivity to such stain related markers.

It should be noted that the amount of additive noise is stain dependent: a standard deviation, $\sigma$, of $0.3$ produces realistic CD68, but a noisy CD34, output.
As the translation process hides non-overlapping inter-stain information, the intermediate stain likely determines which information is encoded.

\section{Self Adversarial Attack Augmentation}
\label{sec:augmentation}

CyleGANs are unsupervised and unpaired, therefore training them does not require additional annotation effort but does require additional stain samples. PAS is a routine stain so these should be readily available. The fact that intermediate representations contain imperceptible noise related to stain features can be used to increase the variance of existing datasets by randomly perturbing the noise. CycleGAN is incapable of performing geometrical changes \cite{CycleGAN2017,gadermayr2018which}, so cannot change the morphological structures in these images, e.g.\ it will not remove glomeruli. Thus, it is safe to use as an augmentation technique in supervised problems related to morphologically consistent structures, in this case glomeruli segmentation.

The proposed augmentation process is described in Fig.\ \ref{fig:aug_approach}. Let us denote PAS as $A$ and an immunohistochemical stain as $B$. During supervised training of a model on $B$ (e.g.\ for glomeruli segmentation), each sample $b_i$ is first translated to PAS, $A'$, using the trained CycleGAN generator $G_{BA}$, with a probability of $50\%$. Next, zero-mean Gaussian noise with standard deviation $\sigma$ is added to the intermediate translation, which is translated back to $B$ using $G_{AB}$, where $\sigma \in (0, \epsilon_{\text{stain}}]$ with uniform probability. The value $\epsilon_{\text{stain}}$ is determined for each staining separately. As such, the input is altered by the arbitrary appearance of stain related markers and the supervised model is forced to be less sensitive to their appearance.

The U-Net \cite{Ronneberger15} gives state-of-the-art performance in glomeruli segmentation \cite{lampert2019strategies} and is adopted herein. The architecture and training details are the same as in \cite{lampert2019strategies}.

\subsection{Dataset}

The U-Net training set comprised all glomeruli from the $4$ training patients - $529$ for CD68 and $568$ for CD34 - and $3685$ and $3958$ tissue patches respectively (to account for the variance of non-glomeruli tissue). The validation sets (2 patients) were composed of $524$ and $598$ glomeruli patches, and $3650$ and $4168$ negative patches for CD68 and CD34 respectively. Patches are standardised to $[0,1]$ and normalised by the mean and standard deviation of the training set. To evaluate the augmentation's effect with few data samples, each training set is split into $5$ folds containing $10\%$, $30\%$, and $60\%$ of each class taken at random. A separate random $10\%$ subset of the training data is extracted to choose $\epsilon_{\text{stain}}$. All models are trained for $250$ epochs, the best performing model on the validation partition is kept, and tested on the $4$ held-out test patients. The average F$_1$-score and standard deviation is reported.

\subsection{Choosing the Level of Noise}

As with all augmentation techniques, a parameter value must be chosen. In this case it is the noise level $\epsilon_\text{stain}$. Since the problem being addressed is supervised, $\epsilon_\text{stain}$ can be optimised experimentally, however, it could be chosen by manually validating the reconstructions. A grid search was conducted on a separate dataset partition containing a random $10\%$ subset of each class. The range $\epsilon_\text{stain} \in [0.01, 0.05, 0.1, 0.3, 0.5, 0.9]$ was tested by averaging $3$ repetitions. It was found that adding noise in the range that produces realistic output improves upon the baseline ($\epsilon_\text{CD68} \le 0.3$ and $\epsilon_\text{CD34} \le 0.1$), confirming that the parameter can be chosen manually. Nevertheless, the best value should be determined for each stain to maximise $F_1$ score and these were found to be $\epsilon_\text{stain} = 0.05$.

\subsection{Results}

Table \ref{tab: results} presents the baseline and noise augmented results with varying amounts of data. The proposed augmentation improves $F_1$ scores unanimously due to increased precision. Recall does not improve since no new task-specific information is added, e.g.\ glomeruli shape or positional variance. Since stain related markers are not indicative of glomeruli in general, the model should largly ignore them. However, fibrotic and sclerotic glomeruli are present, to which the model can wrongly associate a specific pattern or marker. 
For example, fibrotic changes are associated with CD68 positive macrophages \cite{Adler20} and a loss of CD34 positive vascular structures.
Overemphasising immunohistochemical variations via augmentation biases the model to other properties, decreasing recall but disproportionately increasing precision.
\color{black}

\section{Conclusion}
\label{sec:conclusions}

This article studies CycleGAN self-adversarial attacks in translating immunohistochemical stainings to PAS. It presents evidence that imperceptible noise induced by cycle consistency relates to immunohistochemical markers. Perturbing this hidden information causes these markers to appear in different, plausible locations although their physical meaning remains an open question. This finding is used in an augmentation method to increase segmentation accuracy by reducing false positive rates and therefore increasing $F_1$ scores. We also found that the translations result in rich and realistic images, which may provide cellular information and future work will take this direction by investigating their physical meaning, in addition to analysing different reference stains. 

\section{Compliance with Ethical Standards}
\label{sec:ethics}
This study was performed in line with the principles of the Declaration of Helsinki. Approval was granted by the Ethics Committee of Hannover Medical School (Date 12/07/2015, No.\ 2968-2015).

\section{Acknowledgments}
\label{sec:acknowledgments}
This work was supported by: ERACoSysMed and e:Med initiatives by the German Ministry of Research and Education (BMBF); SysMIFTA (project management PTJ, FKZ 031L-0085A; Agence National de la Recherche, ANR, project number ANR-15—CMED-0004); SYSIMIT (project management DLR, FKZ 01ZX1608A); and the French Government through co-tutelle PhD funding.
We thank Nvidia Corporation for donating a Quadro P6000 GPU and the \emph{Centre de Calcul de l'Université de Strasbourg} for access to the GPUs used for this research.
We also thank the MHH team for providing high-quality images and annotations, specifically Nicole Kroenke for excellent technical assistance, Nadine Schaadt for image management and quality control, and Valery Volk and Jessica Schmitz for annotations under the supervision of domain experts.

\bibliographystyle{IEEEbib}
\bibliography{isbi_2021}

\begin{thebibliography}{10}

\bibitem{grote2018crowdsourcing}
A.\ Grote et~al.,
\newblock ``Crowdsourcing of histological image labeling and object delineation
  by medical students,''
\newblock {\em IEEE Trans Med Imaging}, vol. 38, pp. 1284--1294, 2018.

\bibitem{Leo2016EvaluatingSO}
P.\ Leo et~al.,
\newblock ``Evaluating stability of histomorphometric features across scanner
  and staining variations: prostate cancer diagnosis from whole slide images,''
\newblock {\em J Med Imaging}, vol. 3, no. 4, 2016.

\bibitem{csurka2017comprehensive}
G.\ Csurka,
\newblock ``A comprehensive survey on domain adaptation for visual
  applications,''
\newblock in {\em Domain adaptation in computer vision applications},
  chapter~1, pp. 1--35. Springer, 2017.

\bibitem{goodfellow2014generative}
I.\ Goodfellow et~al.,
\newblock ``Generative adversarial nets,''
\newblock in {\em NIPS}, 2014, pp. 2672--2680.

\bibitem{Tellez2019QuantifyingTE}
D.\ Tellez et~al.,
\newblock ``Quantifying the effects of data augmentation and stain color
  normalization in convolutional neural networks for computational pathology,''
\newblock {\em Med Image Anal}, vol. 58, pp. 101544, 2019.

\bibitem{Mercan2020VirtualSF}
C.\ Mercan et~al.,
\newblock ``Virtual staining for mitosis detection in breast histopathology,''
\newblock in {\em ISBI}, 2020.

\bibitem{Lahiani2018VirtualizationOT}
A.\ Lahiani et~al.,
\newblock ``Virtualization of tissue staining in digital pathology using an
  unsupervised deep learning approach,''
\newblock {\em ECDP}, vol. 11435, 2019.

\bibitem{Vasiljevic20}
J.\ Vasiljević et~al.,
\newblock ``Achieving histopathological stain invariance by unsupervised domain
  augmentation using generative adversarial networks,''
\newblock {\em Under Review}.

\bibitem{PathologyGAN20}
A.\ Quiros et~al.,
\newblock ``Pathologygan: Learning deep representations of cancer tissue,''
\newblock in {\em PMLR}, 2020, vol. 121, pp. 669--695.

\bibitem{CycleGAN2017}
J.-Y.\ Zhu et~al.,
\newblock ``Unpaired image-to-image translation using cycle-consistent
  adversarial networks,''
\newblock in {\em ICCV}, 2017, pp. 2242--2251.

\bibitem{gadermayr2018which}
M.\ Gadermayr et~al.,
\newblock ``Which way round? {A} study on the performance of stain-translation
  for segmenting arbitrarily dyed histological images,''
\newblock 2018, pp. 165--173.

\bibitem{Bashkirova19}
D.\ Bashkirova et~al.,
\newblock ``Adversarial self-defense for cycle-consistent {GAN}s,''
\newblock in {\em NeurIPS}, 2019, pp. 635--645.

\bibitem{Chu2017CycleGANAM}
C.\ Chu et~al.,
\newblock ``{CycleGAN}, a master of steganography,''
\newblock in {\em NIPS workshop on Machine Deception}, 2017.

\bibitem{Cytomine16}
R.~Mar{\'e}e et~al.,
\newblock ``Collaborative analysis of multi-gigapixel imaging data using
  cytomine,''
\newblock {\em Bioinformatics}, vol. 32, no. 9, pp. 1395--1401, 2016.

\bibitem{Ronneberger15}
O.~Ronneberger et~al.,
\newblock ``U-{N}et: Convolutional networks for biomedical image
  segmentation,''
\newblock in {\em MICCAI}, 2015, pp. 234--241.

\bibitem{lampert2019strategies}
T.\ Lampert et~al.,
\newblock ``Strategies for training stain invariant {CNN}s,''
\newblock in {\em ISBI}, 2019, pp. 905--909.

\bibitem{Adler20}
M.\ Adler et~al.,
\newblock ``Principles of cell circuits for tissue repair and fibrosis,''
\newblock {\em iScience}, vol. 23, no. 2, pp. 100841, 2020.

\end{thebibliography}

\end{document}